\documentclass[lettersize,journal]{IEEEtran}
\usepackage{bbding}
\usepackage[switch]{lineno}

\usepackage{graphicx}
\usepackage{amsmath}
\usepackage{epsfig}
\usepackage{colortbl}
\usepackage{multirow}
\usepackage{booktabs}
\usepackage{mwe}
\usepackage{ulem}

\usepackage{pifont}
\usepackage{xcolor}
\usepackage{enumitem}
\usepackage{subcaption}
\usepackage{ulem}
\usepackage{bm}
\usepackage{algorithm}
\usepackage{amssymb}
\usepackage{booktabs}
\usepackage[pagebackref,breaklinks,colorlinks]{hyperref}

\definecolor{topcolor}{rgb}{0.9020, 0.9608,0.9882}
\definecolor{bottomcolor}{rgb}{0.9804,0.94117647,0.91764706}

\definecolor{ugreen}{cmyk}{1,0,1,0.498}
\definecolor{lyyblue}{cmyk}{0.8278,0.3333,0,0.2941}
\definecolor{lyygreen}{cmyk}{0.6813,0,0.725,0.3725}
\definecolor{lyyred}{cmyk}{0,0.8855,0.8767,0.1098}
\definecolor{dblue}{cmyk}{1,0.5487,0,0.5569}
\definecolor{lypurple}{HTML}{e0c2c0}
\definecolor{lygreen}{HTML}{eff67b}
\definecolor{lyblue}{HTML}{d5ddef}
\definecolor{lyyellow}{HTML}{fdfab5}
\definecolor{lypink}{HTML}{ffe0db}
\definecolor{lyred}{HTML}{b71a3b}
\definecolor{lygrey}{HTML}{c4c0c2}
\definecolor{lyorange}{HTML}{eab586}
\definecolor{lightgrey}{gray}{0.90}
\usepackage[vlined,linesnumbered,ruled,algo2e]{algorithm2e}
\usepackage[linesnumbered,ruled,vlined]{algorithm2e}
\usepackage[capitalize]{cleveref}
\usepackage{cite}
\begin{document}

\title{CLASH: Complementary Learning with Neural Architecture Search for Gait Recognition}

\author{Huanzhang Dou$^*$, Pengyi Zhang, Yuhan Zhao, Lu Jin, Xi Li
\thanks{$^*$Work done during a research internship at Ant Group.}
\thanks{Huanzhang Dou and Pengyi Zhang are with College of Computer Science, Zhejiang University (email: \{hzdou, pyzhang\} @zju.edu.cn)}
\thanks{Yuhan Zhao is with School of Software Technology, Zhejiang University (email: yuhanzhao@zju.edu.cn) }
\thanks{Lu Jin is with Ant Group. (email: lyla.jl@antgroup.com) }
\thanks{Xi Li is with College of Computer Science and Technology, Zhejiang University. (email: xilizju@zju.edu.cn)}}

\markboth{Journal of \LaTeX\ Class Files,~Vol.~14, No.~8, August~2021}%
{Shell \MakeLowercase{\textit{et al.}}: A Sample Article Using IEEEtran.cls for IEEE Journals}


\maketitle

\begin{abstract}
Gait recognition, which aims at identifying individuals by their walking patterns, has achieved great success based on silhouette. The binary silhouette sequence encodes the walking pattern within the sparse boundary representation. Therefore, most pixels in the silhouette are \textit{under-sensitive} to the walking pattern since the \textit{sparse boundary} lacks \textit{dense} spatial-temporal information, which is suitable to be represented with dense texture. To enhance the sensitivity to the walking pattern while maintaining the robustness of recognition, we present a \textbf{C}omplementary \textbf{L}earning with neural \textbf{A}rchitecture \textbf{S}earc\textbf{H} (CLASH) framework, consisting of walking pattern sensitive gait descriptor named dense spatial-temporal field (DSTF) and neural architecture search based complementary learning (NCL). Specifically, DSTF transforms the representation from the sparse binary boundary into the dense distance-based texture, which is sensitive to the walking pattern at the pixel level. Further, NCL presents a task-specific search space for complementary learning, which mutually complements the sensitivity of DSTF and the robustness of the silhouette to represent the walking pattern effectively. Extensive experiments demonstrate the effectiveness of the proposed methods under both in-the-lab and in-the-wild scenarios. On CASIA-B, we achieve rank-1 accuracy of 98.8\%, 96.5\%, and 89.3\% under three conditions. On OU-MVLP, we achieve rank-1 accuracy of 91.9\%. Under the latest in-the-wild datasets, we outperform the latest silhouette-based methods by 16.3\% and 19.7\% on Gait3D and GREW, respectively.
\end{abstract}

\section{Introduction}
Gait, the walking pattern of individuals, is one of the most promising biometrics since it can be recognized at a long distance without the explicit cooperation of humans. Recently, appearance-based
gait recognition with convolutional neural network~\cite{Huang_2021_ICCV,choi2019user,Lin_2021_ICCV}, which extracts the walking pattern from silhouette sequences, has drawn increasing research attention and revealed great application potentials, \textit{e.g.,} security check~\cite{ Meng_Fu_Yan_Liang_Zhou_Zhu_Ma_Liu_Yang_2020}, video retrieval~\cite{bouchrika2018survey}, and identity identification~\cite{macoveciuc2019forensic}. Specifically, the walking pattern is mainly composed of the body appearance and its motion pattern~\cite{1561189,chao2019gaitset,Fan_2020_CVPR}, \textit{i.e.,} spatial information and temporal information.

 \begin{figure}[t]
    \begin{center}
       \includegraphics[width=0.45\textwidth]{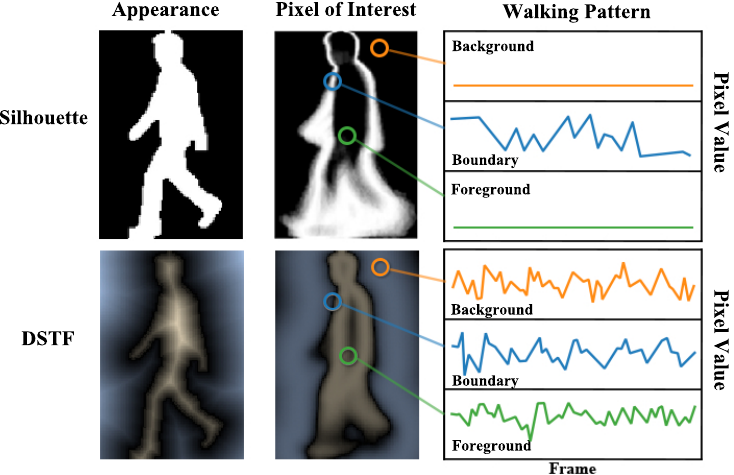}
    \end{center}
    \caption{Comparison between the silhouette and proposed DSTF. The pixel of interest computed by GEnI~\cite{5522296} refers to the set of pixels that contribute most to the walking pattern. The walking patterns of the pixels on the foreground, boundary, and background are reflected in the value changes of the corresponding pixels. The pixel on DSTF with blue denotes the negative pixel value.}
    \label{fig:shoutu}
 \end{figure}

However, the sparse binary boundary representation of the silhouette is under-sensitive to the walking pattern, which means the silhouette cannot informatively represent gait information. First, in the spatial dimension, the appearance of the human body is determined by the sparse junction between the foreground and the background~\cite{wang2002gait,wang2010chrono}, \textit{i.e.,} boundary. However, the boundary is only a small part of the silhouette. Therefore, the boundary representation lacks dense spatial information to fully represent the body appearance. Second, in the temporal dimension, motion changes are sparse and subtle in the whole silhouette because the vast majority of pixels on silhouette do not change over time, resulting in a lack of temporal information. The above two lacks may harm the effectiveness of the walking pattern representation.
Further, the convolutional neural network (CNN) is designed for dense data with \textit{texture-based} representation~\cite{8490955,8697107,geirhos2018imagenet}. Thus, the sparse binary boundary representation increases the difficulty of feature extraction. To intuitively show the sparsity of the boundary representation in spatial/temporal dimension, we visualize the silhouette and utilize GEnI~\cite{5522296} to discover the pixel of interest of the silhouette, which is the set of pixels that contribute most to the walking pattern. As shown in the first-row of~\cref{fig:shoutu}, the walking pattern clues of the boundary representation are too subtle to be captured, \textit{i.e.,} only limited pixels near the boundary could depict the walking pattern while most of the pixels are uninformative and under-sensitive. 


On the contrary, the sparse gait descriptor, \textit{i.e.,} silhouette, has the advantage of being a regularization method to avoid being over-sensitive~\cite{li2017weighted, shao2018spatial}, which is also crucial because the over-fitting problem in gait recognition is relatively common~\cite{shen2022gait}. The preprocessing process of silhouette, such as segmentation, inevitably introduces noise into the walking pattern. Therefore, dense representation could increase the sensitivity to walking patterns but also inevitably increase the sensitivity to noise simultaneously. Although the silhouette cannot informatively represent the walking pattern, it is relatively robust to the noise due to the sparsity. Therefore, modeling the walking pattern sensitive texture-based representation, while utilizing the regularization impact of boundary-based representation to achieve informative and robust representation is a fundamental problem for gait recognition.

Motivated by this, we present a \textbf{C}omplementary \textbf{L}earning with neural \textbf{A}rchitecture \textbf{S}earc\textbf{H} (CLASH) framework, which is composed of walking pattern sensitive gait descriptor named dense spatial-temporal field (DSTF) and neural architecture search based complementary learning (NCL). For DSTF, first, we propose to model the relation between each pixel and the boundary to densely represent spatial information, which is achieved by transforming the representation from the sparse binary boundary into the dense texture, \textit{i.e.,} pair-wise distance between each pixel and its nearest pixel on the boundary. Besides, as shown in~\cref{fig:shoutu}, distance-based representation is also more informative than the sparse representation in the temporal dimension, indicating that distance-based representation could capture subtle motion changes at the pixel level.
Second, considering that the pixel distribution and semantics are different between foreground and background, we perform the foreground/background separation strategy to distinguish them explicitly. As the second row is shown in~\cref{fig:shoutu}, the pixel of interest of DSTF is densely distributed among the whole map and DSTF is more sensitive to the walking pattern than the silhouette. Besides, DSTF has more than four times as much information as a silhouette in terms of Image Entropy~\cite{e21030244}, indicating DSTF could informatively describe the walking pattern.

Then, we propose to perform complementary learning~\cite{Zhang_2018_CVPR,wang2019adaptive,GE2016112,he2020rstaple}, which leverages the sensitivity of DSTF to the walking pattern and the robustness of silhouette to the noise, to achieve informative and robust gait representation, respectively. However, adaptively finding a balance between over-sensitive and under-sensitive from two heterogeneous descriptors is rather difficult, requiring elaborate architecture design and the substantial heuristic effort of human experts through inefficient trial and error. Therefore, to efficiently take full advantage of two heterogeneous and complementary descriptors, we leverage neural architecture search (NAS) to perform complementary learning. Specifically, we design a task-specific search space for complementary learning with a bilevel optimization and we design a multi-descriptor (MD) cell to integrate the features of silhouette and DSTF.

The main contributions can be summarized as follows:
\begin{itemize}
    \item We propose a walking pattern sensitive gait descriptor named dense spatial-temporal field (DSTF). DSTF transforms the representation from the sparse binary boundary into the dense distance-based texture, which could represent the walking pattern informatively, either in the spatial dimension or the temporal dimension.
    
    \item We propose NAS-based complementary learning, which aims to leverage the complementarity between DSTF and silhouette to achieve informative and robust representation. NAS automates the architecture design of complementary learning to fit the property of gait descriptors.
    
    \item Extensive experiments conducted on popular datasets~\cite{1699873,takemura2018multi,zhu2021gait,Zheng_2022_CVPR} demonstrate the effectiveness of the proposed methods under both in-the-lab and in-the-wild scenarios.
\end{itemize}

\label{sec:intro}

\section{Related Work}
\subsection{Gait Recognition}

\noindent\textbf{Model-based Methods.} These methods~\cite{9879229,liao2020model, kastaniotis2016pose,ariyanto2011model,nordin2016survey,benouis2016gait,li2020model} aim at modeling the underlying structure of the human body and extracting walking pattern features for recognition. For instance, Wang \textit{et al.}~\cite{wang2004fusion} utilize the angle change of joints to recognize different individuals. Liao~\textit{et. al.}~\cite{liao2017pose} propose a pose-based temporal-spatial network (PTSN) to extract the temporal-spatial features. PoseMapGait~\cite{LIAO2022514} not only preserves richer cues of the human body compared with the skeleton-based feature but also keeps the advantage of being less sensitive to human shape than the silhouette-based feature. Human mesh~\cite{li2020end,9773349} is also introduced to represent the human body. Further, graph modeling~\cite{9506717,teepe2021gaitgraph} is utilized to extract the relation of the joints. Besides, skeleton data can also be combined with silhouette sequences for better performance than using one gait descriptor~\cite{10096986}. The advantage of these approaches is that they are robust to covariates. However, they rely heavily on accurate detection of the joints and are sensitive to occlusions. Besides, the off-the-shelf pose estimator introduces extra computational overhead.

\noindent\textbf{Appearance-based Methods.} These methods~\cite{Huang_2021_ICCV2,Lin_2021_ICCV,dou2021versatilegait, 10.1145/3343031.3351018,10.1145/3343031.3351018,dou2022metagait,Fan_2023_CVPR,dou2023gaitgci,zhang2023large} directly extract the walking pattern features from silhouette sequences without explicitly modeling the structure of human body. For example, GaitSet~\cite{chao2019gaitset} regards each silhouette sequence as an unordered set. GaitPart~\cite{Fan_2020_CVPR} and Wu \textit{et al.}~\cite{wu2020condition} utilize 1D convolutions to extract temporal clues. GaitGL~\cite{Lin_2021_ICCV} proposes to jointly leverage global and local representation to model the walking pattern. GaitEdge~\cite{liang2022gaitedge} directly extracts the walking pattern from RGB sequences. Besides, there are some works~\cite{wang2010chrono,guha2010differential,el2014new,boulgouris2007gait,el2014new,wang2002gait,wang2003silhouette} that extract the features of the individual only from the boundary of the human body, \textit{e.g.,} Wang \textit{et al.}~\cite{wang2003automatic} utilize the shape analysis to extract gait signatures from the boundary pixels. Appearance-based approaches are popular due to their flexibility, conciseness, and effectiveness. This paper is in the scope of the appearance-based gait recognition method, and we evaluate the proposed framework under both in-the-lab and in-the-wild scenarios.

\begin{figure*}[htbp]
    \begin{center}
       \includegraphics[width=0.96\textwidth]{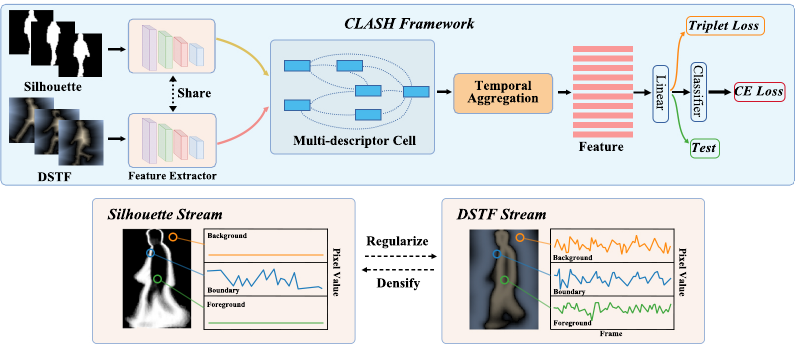}
    \end{center}
    \caption{Overview of CLASH framework. \colorbox{topcolor}{\textit{Top:}} The features of the silhouette and DSTF are extracted by the feature extractor. Then, complementary learning for two heterogeneous descriptors is conducted through neural architecture search, \textit{i.e.,} the multi-descriptor (MD) cell. The final feature is obtained by the temporal aggregation and a linear~\cite{chao2019gaitset}. \colorbox{bottomcolor}{\textit{Bottom:}} The features of silhouette stream and DSTF stream mutually complement each other, \textit{i.e.,} regularize and densify. Note that the arrow between the DSTF and silhouette refers to the interaction rather than the transformation to each other.}
    \label{fig:overview}
 \end{figure*}

\noindent\textbf{Gait Descriptor} Apart from the silhouettes and skeleton, there are various descriptors to represent the walking pattern. Gait energy image~\cite{1561189} is proposed to fuse all frames into a single frame. Chrono-gait image~\cite{wang2010chrono} encodes temporal information with additional colors. Frame difference energy image~\cite{chen2009frame} could suppress the influence of silhouette incompleteness. Optical flow~\cite{8053503, inproceedings123, 6377914} is leveraged to represent the motion pattern. In a similar task like human action recognition, action is represented with a gradient tube via Poisson equation~\cite{gorelick2007actions}. The most relevant method to our work is~\cite{31025e5d8575427c921738f66ef6c50e}, which uses distance to construct a skeleton representation.

Compared to the above-mentioned approaches, the proposed DSTF has several corresponding advantages as follows. First, DSTF is in the form of a sequence rather than a single frame, avoiding information loss. Second, DSTF could be obtained by concise and intuitive distance transform from existing silhouette, instead of introducing extra modality (\textit{e.g.,} 3D mesh) or network (\textit{e.g.,} pose estimation). Besides, distance-based representation is globally sensitive to the boundary at the pixel level, while optical flow is locally sensitive due to the data form and its frame rate.  Third, inspired by the signed distance in the rendering of computer graphics, the effective area is extended to both foreground and background. Besides, considering the different semantic and pixel distributions,  we perform a signed mask and normalization to explicitly separate them and avoid numerical issues, respectively.


\subsection{Neural Architecture Search} Neural architecture search (NAS) aims at automating the network design process. Most NAS approaches search CNN architecture on a small proxy task and transfer the searched CNN to another large target task. Early works utilize either reinforcement learning~\cite{zoph2016neural,baker2016designing} or evolution algorithms~\cite{real2019regularized}. Although achieving considerable performance, they suffer from time-consuming training periods. Then, with the rise of one-shot NAS methods~\cite{bender2018understanding, brock2018smash}, this problem has been greatly solved through training a hyper-network from which each sub-network can reuse its weights. DARTS~\cite{liu2018darts} is the pioneering work for gradient-based NAS, which relaxes the discrete search space to search neural networks in a differentiable way. DetNAS~\cite{chen2019detnas} use NAS for the design of better backbones for object detection. After that, NAS has been widely used in many tasks, such as semantic segmentation~\cite{Lin_2020_CVPR} and tracking~\cite{yan2021lighttrack}. Further, the relevant field such as person ReID~\cite{9444559} and action recognition~\cite{pmlr-v157-zhou21a} has explored leveraging NAS to search backbone and hyperparameters, respectively. In this work, we propose to utilize NAS to perform complementary learning for the first time.

\subsection{Complementary Learning} Mainstream complementary learning~\cite{Zhang_2018_CVPR,wang2019adaptive,GE2016112,he2020rstaple,yuan2024semanticmim,su2023language,Su2023ReferringEC,Su_2024_CVPR,dou2024gvdiffgroundedtexttovideogeneration} aims at leveraging the complementarities between several kinds of information towards comprehensive and discriminative representation. In the person re-identification literature~\cite{zhang2022adaptive,ming2022deep}, Qi \textit{et al.}~\cite{qi2021greyreid} use Grey images and RGB images to enhance the generalization ability. Schumann \textit{et al.}~\cite{schumann2017person}  propose to explore the complementarities between ID-specific features and attribute features. Besides, Zhang \textit{et al.}~\cite{Zhang_2018_CVPR} discover the complementarities of different attention regions for weakly supervised object localization. Although previous methods achieve the remarkable ability of feature representation, these methods are all inefficiently manually designed, which requires the substantial effort of human experts. For example, ASTL~\cite{https://doi.org/10.1049/cvi2.12165} fuses the walking pattern features of template-based and sequence-based representation to extract shape and motion information at the same time by carefully designed multimodal cooperative learning modules.

In contrast to the above manually designed complementary learning approaches, we propose to leverage neural architecture search to efficiently search the fusion module, which could perform complementary learning with two heterogeneous and complementary descriptors in a data-driven paradigm.


\label{sec:related}

\section{Method}
In this section, we illustrate our complementary learning with neural architecture search (CLASH) framework, which consists of a dense spatial-temporal field (DSTF) and neural architecture search based complementary learning (NCL). Then, we summarize the whole objective and the two-stage optimization strategy for CLASH at the end of this section.

\subsection{Overview}
The overview of CLASH is presented in~\cref{fig:overview}. First, the weight-sharing feature extractor aims at extracting features for each heterogeneous descriptor. Second, we leverage a multi-descriptor (MD) cell to conduct complementary learning between the features of the silhouettes and DSTF from the feature extractor. Specifically, the architecture of the MD cells is obtained by NAS. Third, temporal aggregation is adopted to aggregate features along the temporal dimension to produce the final features. Finally, we use the combination of triplet loss and cross-entropy loss as the whole objective.
\subsection{Dense Spatial-temporal Field}

To design a walking pattern sensitive gait descriptor, the key is to model a dense texture-based representation that is more suitable for the convolutional neural network to perform feature extraction. In this section, we propose to encode the spatial-temporal information with dense distance-based texture representation and foreground/background separation strategy. Because the distance-based representation is dense and implicitly directional, the proposed gait descriptor is named as dense spatial-temporal field (DSTF).


\noindent\textbf{Dense Distance-based Texture Representation.}
As described in~\cref{sec:intro}, most pixels on the foreground/background of boundary representation cannot represent valid spatial-temporal information. Motivated by this, we propose to model dense spatial-temporal information by transforming the representation from the sparse binary boundary into the dense distance-based texture. Note that we use texture (\textit{i.e.,} distance representation) to refer to low-frequency information that changes slowly, and boundary (\textit{i.e.,} edge) to refer to high-frequency information that significantly changes.
\begin{figure}
\centering
	\subcaptionbox{Silhouette.\label{fig:3a}}{\includegraphics[width = 0.1\textwidth]{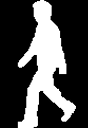}}
	\hspace{0.3cm}
	\subcaptionbox{Fore-DT.\label{fig:3b}}{\includegraphics[width = 0.1\textwidth]{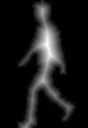}}
	\hspace{0.3cm}
	\subcaptionbox{Back-DT.\label{fig:3c}}{\includegraphics[width = 0.1\textwidth]{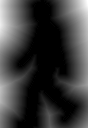}}
	\hspace{0.3cm}
	\subcaptionbox{DSTF.\label{fig:3d}}{\includegraphics[width = 0.1\textwidth]{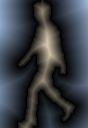}}
\caption{Comparison between silhouette, foreground of Bi-DT (Fore-DT), background of Bi-DT (Back-DT), and a single frame of DSTF. The background is marked with blue for visualization of negative pixel values.}
\label{fig:dtdif}
\end{figure}

\begin{figure}
\centering
	\subcaptionbox{silhouette.}{\includegraphics[width = 0.1\textwidth]{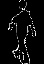}}\hspace{1.5cm}
	\subcaptionbox{DSTF.}{\includegraphics[width = 0.1\textwidth]{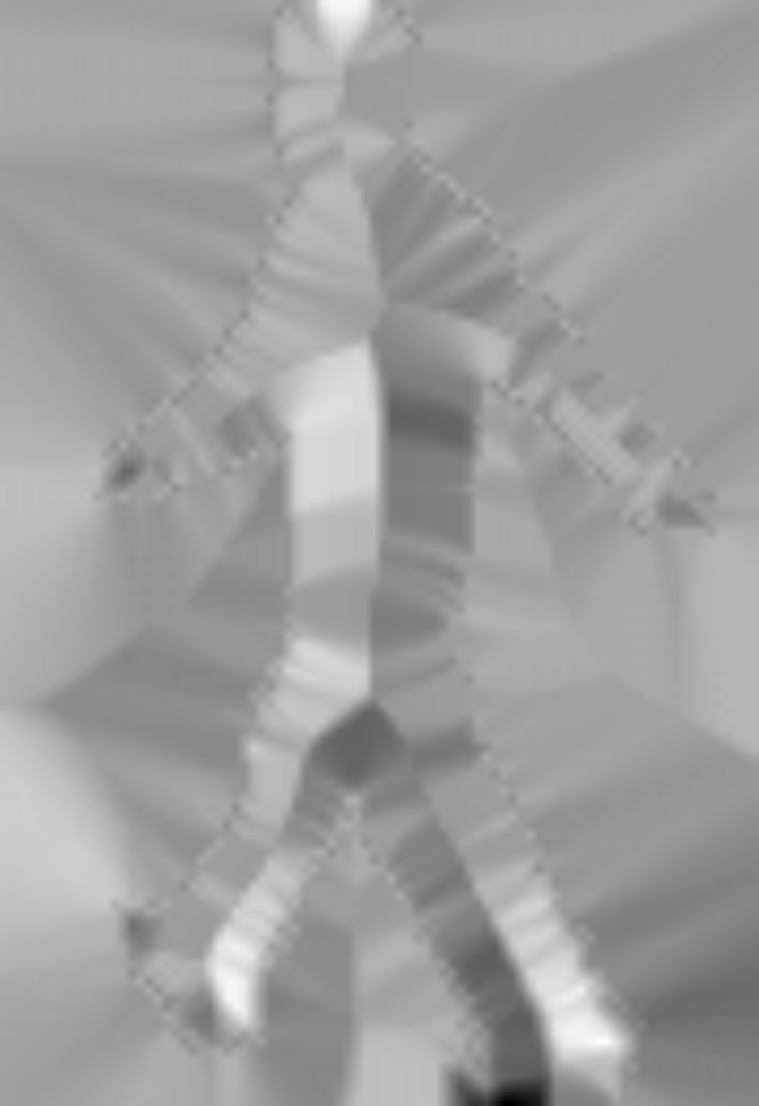}}
	
\caption{Motion information, \textit{i.e.,} frame difference comparison. Most pixels on the single frame of DSTF could change over time to informatively represent the temporal information, while most pixels on silhouette cannot change over time.} 
\label{fig:motiont}
\end{figure}

Technically, we propose Bidirectional Distance Transform (Bi-DT) to encode the gait pattern into the texture representation, converting the values of each foreground/background pixel from binary values into the pair-wise distance between each pixel and its nearest pixel on the boundary. To better understand Bi-DT, the examples of Bi-DT in the foreground (Fore-DT) and background (Back-DT) are shown in~\cref{fig:3b} and ~\cref{fig:3c}, respectively. Bi-DT is inspired by Distance Transform, an image processing algorithm. Specifically, given a silhouette frame $\bm{i}$, which can be divided into two groups: boundary pixels $\bm{P}_{B}$ and other pixels $\bm{P}_{O}$. $\bm{P}_{O}$ is composed of foreground pixels $\bm{P}_{fore}$ and background pixes $\bm{P}_{back}$. The value of each pixel $\bm{p}$ on Bi-DT map $\bm{i}_{bd}$ is formulated as:
\begin{equation}
\bm{i}_{bd}(\bm{p}) =
    \begin{cases}
    \min\limits_{\bm{q} \in \bm{P}_{B}} \bm{\mathrm{Dist}}(\bm{p}, \bm{q}),  & \text{$\bm{p} \in \bm{P}_{O}$}\\
    0,  & \text{$\bm{p} \in \bm{P}_{B}$}
    \end{cases},
\end{equation}
where $\bm{\mathrm{Dist}}(\cdot, \cdot )$ is Euclidean distance function. Note that the boundary in silhouette changes over time and the distance is sensitive to the boundary at the pixel level. Therefore, the distance-based representation could inherently capture the subtle motion changes as shown in~\cref{fig:motiont}.



\noindent\textbf{Foreground/background Separation Strategy.} Considering the different distribution and semantics between foreground and background, we propose to separate them in two aspects explicitly. Inspired by the signed distance field in rendering~\cite{Liu_2020_CVPR}, the foreground and background of $\bm{i}_{dstf}$ are separated by a signed function concisely.


First, the different distance distribution of foreground and background in the silhouette results in the maximum pixel value on $\bm{i}_{dstf}$ concentrating in the corner of the background, while the pixel value of the foreground is relatively small. Therefore, we propose to perform normalization on foreground/background separately to alleviate this numerical issue. Second, the isotropic distance representation cannot explicitly distinguish the foreground/background points, which are about boundary symmetry. However, the semantics of foreground and background are different. Thus, we propose to leverage a signed mask to separate the foreground and background as:
\begin{equation}
      \resizebox{1\linewidth}{!}{$
            \displaystyle
\bm{i}_{dstf}(\bm{p}) = \bm{\mathrm{Sign}}(\bm{p}) * \bm{i}_{bd}(\bm{p}), \,\,\,\, s.t. \,\, \bm{\mathrm{Sign}}(\bm{p})=
    \begin{cases}
    1, & \text{$\bm{p} \in \bm{P}_{fore}$}\\
    -1,  & \text{$\bm{p} \in \bm{P}_{back}$}
    \end{cases}.
        $}
    \label{equ:sign}
\end{equation}

As a result, DSTF $\bm{I}_{dstf}$ is composed of the sequence of $\bm{i}_{dstf}$ and DSTF could densely represent spatial-temporal information as shown in~\cref{fig:shoutu}. After traversing the mainstream dataset, \textit{i.e.,} CASIA-B, OU-MVLP, Gait3D, and GREW, the average Image Entropy~\cite{e21030244} of $\bm{i}_{dstf}$ is 4.03 times that of the silhouette, which indicates that DSTF could informatively represent the walking pattern. Note that we aim at verifying the superiority of dense representations over sparse ones rather than enumerating all the densification techniques, since there are many alternatives. We choose the distance-based representation because it is concise, efficient, and sensitive to the walking pattern at the pixel level.

\subsection{NAS-based Complementary Learning}

It is suboptimal to use existing complementary learning methods because DSTF is a newly designed gait descriptor, and existing methods may not satisfy the property of DSTF. Further, if designing a new architecture to adapt the complementary learning of silhouette and DSTF, a substantial heuristic effort is inevitable. Therefore, we introduce NAS for automating the architecture design of complementary learning in a data-driven paradigm, which is more consistent with the property of the gait descriptor. 

However, simply introducing NAS into gait recognition confronts some problems. First, as a video task, directly leveraging existing image-based NAS methods cannot effectively extract temporal information. Further, improving the adaptiveness of search architecture is necessary to perceive the properties of gait descriptors when performing complementary learning. Second, processing sequential inputs introduces heavy computational overhead. Besides, the complexity of the binary silhouette is lower than RGB data. Therefore, we propose to compress the cell topology of NCL.

In this section, we propose to leverage NAS to search the feature fusion module, which could  excavate the complementarity between the silhouette and DSTF for informative and robust gait feature representation in a data-driven paradigm to reduce human labor. Therefore, the proposed method is named as NAS-based complementary learning (NCL). We illustrate the proposed NCL from two aspects: the task-specific search space and the topology of the proposed cells.

\noindent\textbf{Task-specific Search Space.}  In the NAS literature, search space, which covers all possible candidate architectures to be searched, is crucial for NAS design. However, a standard search space only contains 2D convolutional layers, pooling layers, \textit{etc.} These simple operations can hardly model the spatio-temporal relation between two heterogeneous descriptors~\cite{9454270, Sun_2021_CVPR, 8918443} explicitly, which requires elaborate design. Note that CLASH focuses on exploring a task-specific search space to efficiently search the architecture that is suitable for the complementary learning layer, excluding the feature extractor for saving the computation resource.

 Specifically, we design a multi-descriptor (MD) cell, which leverages the complementarities between texture-based and boundary-based descriptors to achieve walking pattern sensitive representation while avoiding being over-sensitive. The MD cell aggregates the extracted features $\bm{F}_{sil}$ and $\bm{F}_{dstf}$ from the silhouettes and DSTF, respectively.  
\begin{equation}
     \bm{F}_{MD} = \bm{\mathrm{MD}}(\bm{F}_{sil}, \bm{F}_{dstf}),
    \label{eq:mmcell}
\end{equation}
where $\bm{F}_{MD}$ is the output of the MD cell. Then, we utilize GeM pooling~\cite{gu2018attention} along the temporal dimension to adaptively aggregate the temporal information, which is applied to $\bm{F}_{MD}$ to obtain the temporal aggregated features $\bm{F}_{agg}$:

\begin{equation}
     \bm{F}_{agg} = (\bm{\mathrm{Avg}}^{T\times 1\times 1}(\bm{F}_{MD}^{\bm{k}}))^{\frac{1}{\bm{k}}},
    \label{eq:mmcell}
\end{equation}
where $\bm{\mathrm{Avg}}^{T\times 1\times 1}$ is the 3D average pooling with the kernel size of $(T, 1, 1)$. $T$ is the number of the temporal dimension of $\bm{F}_{MD}$ and $\bm{k}$ is a learnable parameter to adaptively control the pooling form. Specifically, when $\bm{k}=1$, GeM pooling is equivalent to average pooling, and when $\bm{k}\to \infty$, GeM pooling is equivalent to max pooling. The finally feature $\bm{F}_{final}$ is acquired by applying separate FC layers~\cite{Lin_2021_ICCV} on $\bm{F}_{agg}$.

\noindent\textbf{Cell Topology.} For simplicity and efficiency, the topology of the MD cell is formulated as shown in~\cref{fig:cell}. This topology is represented by a directed acyclic graph (DAG) consisting of an ordered sequence of $N$ nodes, denoted as $\mathcal{N}=\{\bm{x}^{(1)},\dots, \bm{x}^{(N)}\}$. Each node $\bm{x}^{(i)}$ is a latent representation (\textit{i.e.,} feature map) and each directed edge $(i, j)$ is associated with some candidate operations $\bm{o}^{(i, j)}$ (\textit{e.g.,} conv, pooling) that transform $\bm{x}^{(i)}$. Each intermediate node $\bm{x}^{(j)}$ is computed based on all of its predecessors as:

 \begin{figure}[t]
 \small
    \begin{center}
       \includegraphics[width=0.45\textwidth]{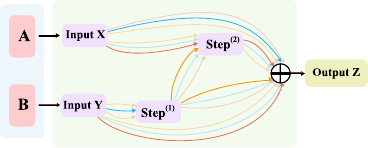}
    \end{center}
    \caption{Illustration of the cell topology. A and B represent the features of the two gait descriptors and are assigned to the two input nodes Input X and Input Y, respectively. The two input nodes, together with two intermediate nodes and one output node, form the whole topology of the cell.}
    \label{fig:cell}
 \end{figure}

\begin{equation}
    \bm{x}^{(j)}=\sum_{i<j}\bm{o}^{(i,j)}(\bm{x}^{(i)}),
    \label{eq:cell}
\end{equation}
then, to make the search space continuous, we relax the categorical choice of a specific operation to softmax over all possible operations:
\begin{equation}
  \widetilde{\bm{o}}^{(i,j)}(\bm{x}^{(i)})=\sum_{\bm{o}\in \mathcal{O}}\frac{\bm{\mathrm{\exp}}{(\bm{\alpha}_o^{(i,j)})}}{\sum_{\bm{o}^{\prime}\in \mathcal{O}}\bm{\mathrm{\exp}}(\bm{\alpha}_{\bm{o}^{\prime}}^{(i,j)})}\bm{o}(\bm{x}^{(i)}),
\end{equation}
where $\mathcal{O}$ is a candidate operation set, and each operation represents some function $\bm{o}(\cdot)$ to be applied to $\bm{x}^{(i)}$. The weights of mixed operation $\widetilde{\bm{o}}^{(i,j)}$  for edge $(i, j$) are parameterized by $\bm{\alpha}_o^{(i, j)}$. The whole searchable architecture is the complementary learning layer, which can be parameterized by $\bm{\alpha}_{MD}$. After architecture searching, a discrete architecture can be obtained by replacing each mixed operation $\widetilde{\bm{o}}^{(i,j)}$ with the most likely operation, \textit{i.e.,} $\bm{o}^{(i,j)}=\bm{\mathrm{argmax}}_{o\in \mathcal{O}}\bm{\alpha}_o^{(i,j)}$.

For the operation set, considering the heterogeneity between two descriptors, more complex operations are needed to improve the network capacity to conduct complementary learning. Besides, the operation set should be diverse to cover the frequently-used operations, which could take advantage of the prior knowledge of network design by human experts. 
Therefore, we add spatial attention, channel attention, temporal attention, and self-attention to the operation set. Further, to extract temporal information effectively, we replace the basic 2D operations with a variety of 3D operations. To summarize, we design a task-specific operation set $\bm{\mathcal{O}}_{gait}$ as follows:

\begin{minipage}{0.55\linewidth}
    \small
    \medbreak
    \begin{itemize}[leftmargin=*, nolistsep]
    \item $3\hspace{-0.3mm}\times3\times3$ depthwise-separable conv
    \item $5\hspace{-0.3mm}\times5\times5$ depthwise-separable conv
    \item $3\hspace{-0.3mm}\times3\times3$ atrous conv with rate $2$
    \item $5\hspace{-0.3mm}\times5\times5$ atrous conv with rate $2$
    \item $3\hspace{-0.3mm}\times3\times3$ average pooling
    \item $3\hspace{-0.3mm}\times3\times3$ max pooling

    \end{itemize}
    \medbreak
    \end{minipage}
    \begin{minipage}{0.39\linewidth}
    \small
    \medbreak
    \begin{itemize}[leftmargin=*, nolistsep]
    
    \item skip connection
    \item no connection (zero)
    \item channel attention
    \item spatial attention
    \item temporal attention
    \item self-attention
    \end{itemize}
    \medbreak
    \label{ops}
    \end{minipage}
\vspace{-0.2cm}

\subsection{Optimization}
In this part, we introduce the objective of CLASH. Then, we illustrate the two-stage optimization process of CLASH, \textit{i.e.,} the architecture search and the framework optimization.

\noindent\textbf{Objective.}
To effectively conduct the architecture search and model training, the objective $\bm{\mathcal{L}}_{total}$ is set to the combination of triplet loss~\cite{hermans2017defense} $\bm{\mathcal{L}}_{tri}$ and cross-entropy loss $\bm{\mathcal{L}}_{ce}$:

\begin{equation}
    \bm{\mathcal{L}}_{total} =  \bm{\mathcal{L}}_{tri} + \bm{\mathcal{L}}_{ce}.
    \label{eq:total}
\end{equation}

\noindent\textbf{Architecture Search.} The first stage aims at searching for an effective architecture $\bm{\alpha}^*$. The architecture parameter $\bm{\alpha}$ and the network weights $\bm{w}$ are jointly optimized with the bilevel optimization~\cite{anandalingam1992hierarchical,colson2007overview}.

\begin{equation} 
    \begin{split}
        \min_{\bm\alpha} \quad & \bm{\mathcal{L}}_{val}(\bm{w}^*(\bm{\alpha}), \bm{\alpha})  \\
        \text{s.t.} \quad &\bm{w}^*(\bm{\alpha}) = \bm{\mathrm{argmin}}_{\bm{w}} \enskip \bm{\mathcal{L}}_{train}(\bm{w}, \bm{\alpha}),
    \end{split}
    \label{eq:search}
  \end{equation}
where $\bm{\mathcal{L}}_{val}$ and $\bm{\mathcal{L}}_{train}$ are validation and training loss during searching, respectively. For simplicity, they are both set to $\bm{\mathcal{L}}_{total}$. We summarize the search process as~\cref{algorithm}:

\begin{algorithm2e}[h]

    \small
    \caption{Search Process of CLASH.\label{algorithm}}
    \SetKwInput{Kwinit}{Init}{}{}
    \SetKwInput{kwInput}{Input}{}{}
    \SetKwRepeat{Do}{while}{do}
    \SetKwInput{kwOutput}{Output}{}{}
    \kwInput{Training set $\bm{\mathcal{D}}_T$; search space $\mathcal{A}$; network weights $\bm{w}$; cell architecture $\bm{\alpha}$; a factor $\bm{u}$ to balance the update frequency of $\bm{\alpha}$ and $\bm{w}$.} 
    \kwOutput{The searched cell architecture $\bm{\alpha}^*$.}
    Split training set $\bm{\mathcal{D}}_T$ into $\bm{\mathcal{D}}_{train}$ and $\bm{\mathcal{D}}_{val}$ to optimize $\bm{w}$ and $\bm{\alpha}$, respectively; \\
    Create a mixed operation $\widetilde{\bm{o}}^{(i,j)}$ parameterized by $\bm{\alpha}^{(i,j)}$ for each edge $(i,j)$;\\
    \While{\rm not converged}
    {

    \For{$\bm{t}$ = \rm{1} \rm{\textbf{to}} $\bm{u}$}
    {Randomly sample a mini-batch $\bm{d}_{train}$$\subseteq$ $\bm{\mathcal{D}}_{train}$;  \\
    
    Update the network weights $\bm{w}$ by descending $\bm{\nabla}_{\bm{w}}\bm{\mathcal{L}}_{train}(\bm{w}, \bm{\alpha})$ on $\bm{d}_{train}$; \\}
    Randomly sample a mini-batch $\bm{d}_{val}\subseteq \bm{\mathcal{D}}_{val}$;\\
    Update the architecture $\bm{\alpha}$ by descending $\bm{\nabla}_{\alpha}\bm{\mathcal{L}}_{val}(\bm{w}^*(\bm{\alpha}), \bm{\alpha}))$ on $\bm{d}_{val}$;\\
        
    }

   Return searched cell architecture $\bm{\alpha}^*$.
   
\end{algorithm2e}

\noindent\textbf{Framework Optimization.}
With the searched cell architecture $\bm{\alpha}^*$, the second stage aims at retraining the the whole framework with the same loss of the searching stage.
\begin{align}
	\bm{w}^* = \min_{\bm{w}} \bm{\mathcal{L}}_{total}(\bm{w}, \bm{\alpha}^*).
\end{align}

\section{Experiments}

\subsection{Datasets and Evaluation Protocols}
We conduct experiments under the in-the-lab scenarios (OU-MVLP~\cite{takemura2018multi} and CASIA-B~\cite{1699873}) and in-the-wid scenarios (Gait3D~\cite{Zheng_2022_CVPR} and GREW~\cite{zhu2021gait}).

\noindent\textbf{OU-MVLP~\cite{takemura2018multi}.} It is the largest in-the-lab gait dataset, which is composed of 10307 subjects (5153 subjects for training and the rest for testing). Each subject contains two groups of videos and 14 views, which are uniformly distributed between [0$^{\circ}$, 90$^{\circ}$] and [180$^{\circ}$, 270$^{\circ}$]. Following the protocol in~\cite{chao2019gaitset}, the sequences with index \#01 are regarded as the gallery, while the rest sequences with index \#02 are regarded as the probe.

\begin{table*}[!ht]
\centering
\renewcommand{\arraystretch}{0.8}
\setlength{\tabcolsep}{1.6mm}
\caption{Rank-1 (\%) performance comparison on OU-MVLP, excluding the identical-view cases.}
\begin{tabular}{l|c|cccccccccccccc|c}
\toprule
\multirow{2}{*}{Method} & \multicolumn{1}{c|}{\multirow{2}{*}{Venue}} & \multicolumn{14}{c|}{Probe View}                                                               & \multirow{2}{*}{Mean} \\ \cmidrule{3-16}
                        & \multicolumn{1}{c|}{}                      & 0$^{\circ}$ & 15$^{\circ}$ & 30$^{\circ}$ & 45$^{\circ}$ & 60$^{\circ}$ & 75$^{\circ}$ & 90$^{\circ}$ & 180$^{\circ}$ & 195$^{\circ}$ & 210$^{\circ}$ & 225$^{\circ}$ & 240$^{\circ}$ & 255$^{\circ}$ & \multicolumn{1}{c|}{270$^{\circ}$} &      \\ \midrule
                        GEINet~\cite{shiraga2016geinet}   & ICB16&11.4&29.1&41.5&45.5&39.5&41.8&38.9&14.9&33.1&43.2&45.6&39.4&40.5&36.3&35.8                       \\
                                            GaitSet~\cite{chao2019gaitset}                                      & AAAI19  &79.5&87.9&89.9&90.2&88.1&88.7&87.8&81.7&86.7&89.0&89.3&87.2&87.8&86.2&87.1           \\
                      GaitPart~\cite{Fan_2020_CVPR}   &CVPR20                                  &   82.6&88.9&90.8&91.0&89.7&89.9&89.5&85.2&88.1&90.0&90.1&89.0&89.1&88.2&88.7    \\
                    GLN~\cite{hou2020gait}           &ECCV20                               &  83.8&90.0&91.0&91.2&90.3&90.0&89.4&85.3&89.1&90.5&90.6&89.6&89.3&88.5&89.2         \\
                    CSTL~\cite{Huang_2021_ICCV}              &ICCV21                           &   87.1&91.0&91.5&91.8&90.6&90.8&90.6&89.4&90.2&90.5&90.7&89.8&90.0&89.4&90.2           \\
                      3DLocal~\cite{Huang_2021_ICCV2}       &ICCV21                               &   86.1&91.2&92.6&\textbf{92.9}&92.2&91.3&91.1&86.9&90.8&\textbf{92.2}&92.3&91.3&91.1&90.2&90.9        \\
                     GaitGL~\cite{Lin_2021_ICCV}        &ICCV21                               &  84.9&90.2&91.1&91.5&91.1&90.8&90.3&88.5&88.6&90.3&90.4&89.6&89.5&88.8&89.7       \\
                     GaitMPL~\cite{Lin_2021_ICCV}        &TIP22                               &  83.9&90.1&91.3&91.5&91.2&90.6&90.1&85.3&89.3&90.7&90.7&90.7&89.8&88.9&89.6     \\
                     Lagrange~\cite{chai2022lagrange}   &  CVPR22                                          & 85.9  & 90.6    & 91.3   & 91.5   & 91.2   & 91.0   &90.6    &88.9     &89.2     &90.5     &90.6     & 89.9    & 89.8    &     89.2                     & 90.0    \\ 
                     \textbf{Ours}                                         & -- &\textbf{91.0}&\textbf{92.2}&\textbf{92.3}&92.6&\textbf{92.7}&\textbf{92.0}&\textbf{92.0}&\textbf{91.8}&\textbf{91.6}&91.6&\textbf{92.5}&\textbf{92.1}&\textbf{91.2}&\textbf{91.3}&\textbf{91.9}    \\ \bottomrule

\end{tabular}
 \label{tab:ou}
\end{table*}
\noindent\textbf{CASIA-B~\cite{1699873}.} It contains 124 subjects and 11 uniformly distributed views. Each view contains ten sequences with three walking conditions, \textit{i.e.,} 6 NM (normal), 2 BG (carrying), and 2 CL (clothing). The protocol is adopted as~\cite{chao2019gaitset}: small-scale training (ST), medium-scale training (MT), and large-scale training (LT). For these three settings, the first 24/62/74 subjects are chosen as the training set, and the rest 100/62/50 subjects are used for the test, respectively. During the test, the first 4 sequences of the NM condition of each subject are used as the gallery, and the rest are used as the probe.

\noindent\textbf{Gait3D~\cite{Zheng_2022_CVPR}.} Gait3D is the latest in-the-wild dataset, which contains 4000 subjects and 25309 sequences collected in an unconstrained supermarket. Following the official protocol~\cite{zhu2021gait}, 3000 subjects and the rest 1000 subjects are chosen as the training set and the test set, respectively. During the evaluation, one sequence of each subject is deemed as the query, and the rest becomes the gallery. Compared to the in-the-lab datasets, Gait3D is challenging due to the 3D viewpoints, irregular walking speed, occlusion, \textit{etc.} 

\noindent\textbf{GREW~\cite{zhu2021gait}.} GREW is one of the largest in-the-wild datasets, consisting of 26345 subjects collected from 882 cameras. It contains 4 modalities: silhouette, optical flow, 2D pose, and 3D pose. Following the protocol in~\cite{zhu2021gait}, 20000 subjects are for training, and 6000 subjects are for testing. During the evaluation, 2 sequences of each subject are regarded as the probe and the other 2 sequences are the gallery.

\subsection{Implementation Details}

\noindent\textbf{Network Configurations.}  The feature extractor of CLASH is four layers 3D convolutional network with the channel (32, 64, 128, 128) on CASIA-B and add extra 2 layers with doubled channels for other datasets. The feature extractor using different descriptors shares the network parameters.

\begin{table}[t]
    \centering
    \caption{Comparison with prevailing methods of rank-1~(\%) and parameters at the resolution of $64\times 44$ on CASIA-B.}
    \renewcommand{\arraystretch}{0.8}
   
    \begin{tabular}{l|c|ccccc}
    \toprule
    Method & Venue & NM   & BG   & CL   & Mean & Param. \\ 
    \midrule
      GaitSet~\cite{chao2019gaitset}  & AAAI19 & 95.0 & 87.2 & 70.4 & 84.2  &2.59\\ 
  GaitPart~\cite{Fan_2020_CVPR} & CVPR20 & 96.2 & 91.5 & 78.7 & 88.8 &1.20\\ 
    GLN~\cite{hou2020gait}      & ECCV20 & 96.9 & 94.0 & 77.5 & 89.5 &14.70\\ 
   MT3D~\cite{lin2020gait}     & MM20   & 96.7 & 93.0 & 81.5 & 90.4& 3.20\\ 
   DynamicGait~\cite{wu2020condition} &TIP21 & 94.8 & 88.8 & 81.8 & 88.5&-- \\
    CSTL~\cite{Huang_2021_ICCV}     & ICCV21 & 97.8 & 93.6 & 84.2 & 91.9&9.09 \\ 
   3DLocal~\cite{Huang_2021_ICCV2}  & ICCV21 & 97.5 & 94.3 & 83.7 & 91.8&4.26\\ 
    GaitGL~\cite{Lin_2021_ICCV}   & ICCV21 & 97.4 & 94.5 & 83.6 & 91.8 &2.49\\ 
    GaitMPL~\cite{9769988}& TIP22& 95.5 & 92.9 & 87.9 & 92.1&--\\
    Lagrange~\cite{chai2022lagrange}        & CVPR22     &   96.9    &  93.5  &   86.5   &   92.3 &  -- \\
    \textbf{Ours}     &  -      &  \textbf{98.3}    &\textbf{95.3}      &  \textbf{88.0}   &  \textbf{93.9} & 3.51 \\
    \bottomrule
    \end{tabular}
    \label{tab:avg}
    \end{table}

\begin{table}[t]
    \centering
    \caption{Comparison with prevailing methods of rank-1~(\%) at the resolution of $128\times 88$ on CASIA-B.}
    \renewcommand{\arraystretch}{0.8}
    \begin{tabular}{l|c|cccc}
    \toprule
    Method & Venue & NM   & BG   & CL   & Mean   \\ 
    \midrule
     GaitSet~\cite{chao2019gaitset}  & AAAI19 & 95.6 & 91.5 & 75.3 & 87.5 \\ 
    GLN~\cite{hou2020gait}      & ECCV20 & 96.9 & 94.0 & 77.5 & 89.5 \\ 
    CSTL~\cite{Huang_2021_ICCV}     & ICCV21 & 98.0 & 95.4 & 87.0 & 93.4 \\ 
    3DLocal~\cite{Huang_2021_ICCV2}  & ICCV21 & 98.3 & 95.5 & 84.5 & 92.7\\ 
   
    \textbf{Ours}     &  -      &  \textbf{98.8}    &\textbf{96.5}      &  \textbf{89.3}   &  \textbf{94.8}   \\
    \bottomrule
    \end{tabular}
    \label{tab:avg128}
    \end{table}



\noindent\textbf{Search Configurations.} Considering the efficiency of searching,  the MD cell is composed of two input nodes, two intermediate nodes, and one output node. We randomly select 50\% sequences of each ID from training set $\bm{\mathcal{D}}_{T}$ as $\bm{\mathcal{D}}_{train}$ and the other sequences as $\mathcal{D}_{val}$, which are used to optimize parameter $\bm{w}$ and architecture $\bm{\alpha}$, respectively. Then, we use the same resolution as the training stage, a batch size of (8, 2) for $P\times K$ sampling, and 60k iterations for CASIA-B, where $P$ and $K$ are the number of subjects and the number of sequences of each ID when sampling. The batch size is (32, 4) for other datasets. The architecture $\alpha$ is optimized by Adam, with an initial learning rate $1.0^{-4}$ and the hyperparameter of Adam $\beta=(0.5, 0.999)$. The network weights are also optimized by Adam but use $\beta=(0.9,0.999)$. For simplicity, we adopt the same objective for $\bm{\mathcal{L}}_{train}$ and $\bm{\mathcal{L}}_{val}$. The margin is 0.2 in triplet loss. The balance factor $u$ in~\cref{algorithm}  is set to 1. The search process is conducted in the complementary learning layer, only taking 5 hours, which is more efficient than human experts' heuristic efforts. 

\noindent\textbf{Training Configurations.} During training, we resize each frame to $64\times 44$ or $128\times 88$ following~
\cite{hou2020gait, Huang_2021_ICCV2}. In a mini-batch, the batch size is set to (8, 8) for CASIA-B and (32, 8) for other datasets.  Each sequence is sampled according to~\cite{Fan_2020_CVPR}. Adam optimizer is used with a learning rate of $1.0^{-4}$. The margin is set as 0.2 in triplet loss. In CASIA-B, the model is trained for 80k iterations. The iteration of OU-MVLP, Gait3D, and GREW is 200k, 150k, and 200k, respectively. 

\begin{figure}
\centering
	\subcaptionbox{NM}{\includegraphics[width = 0.2\textwidth]{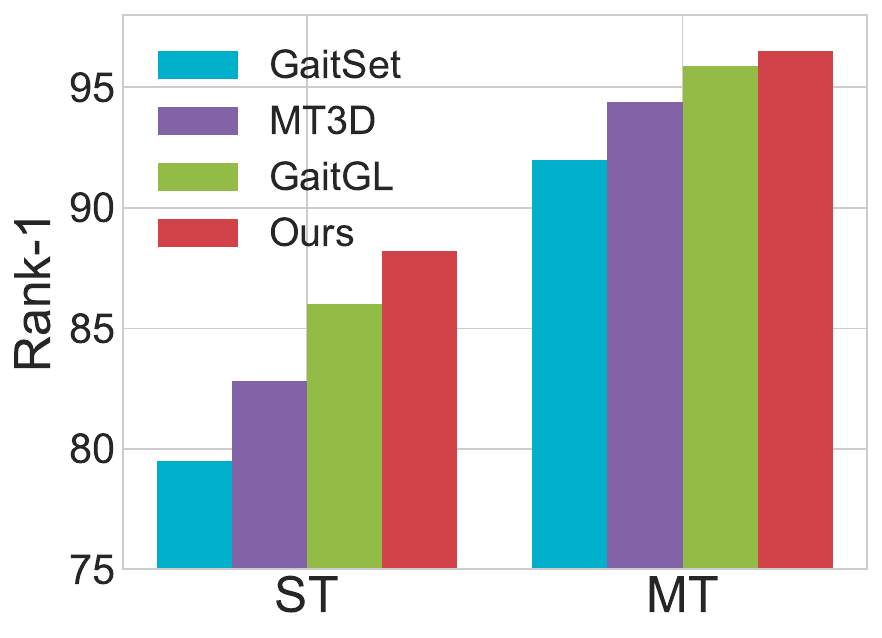}}
	\hspace{0.07cm}
	\subcaptionbox{BG}{\includegraphics[width = 0.2\textwidth]{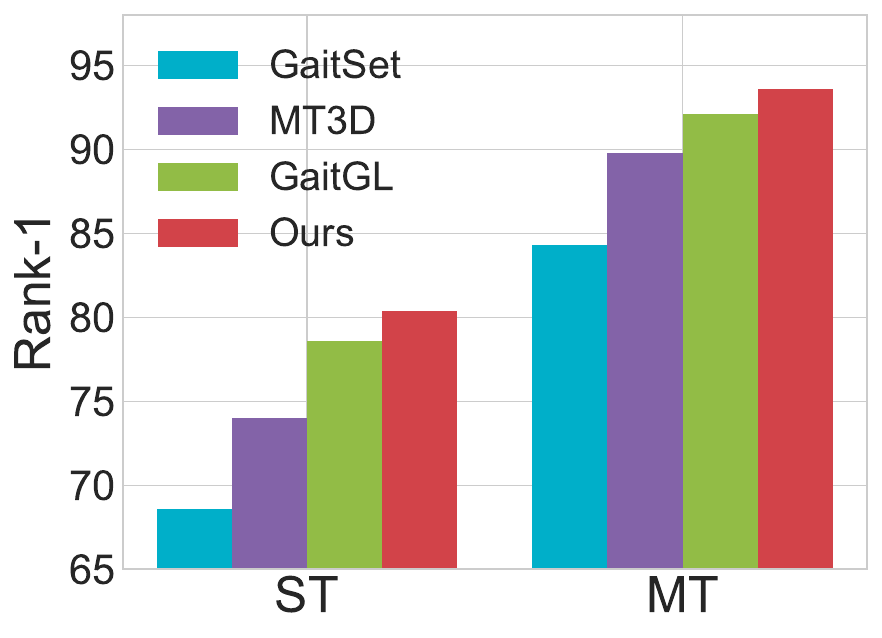}}
	\subcaptionbox{CL}{\includegraphics[width = 0.2\textwidth]{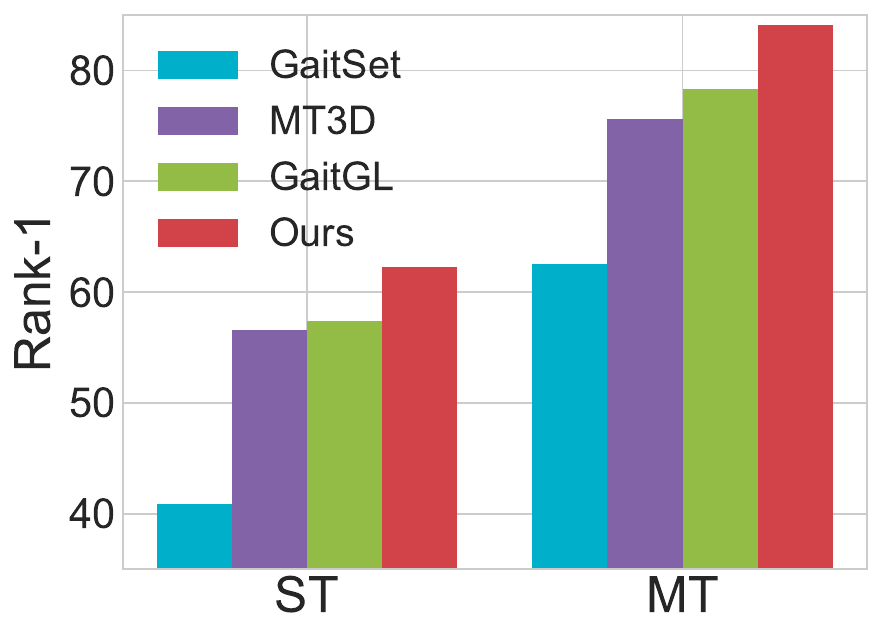}}
	\hspace{0.07cm}
	\subcaptionbox{Mean}{\includegraphics[width = 0.2\textwidth]{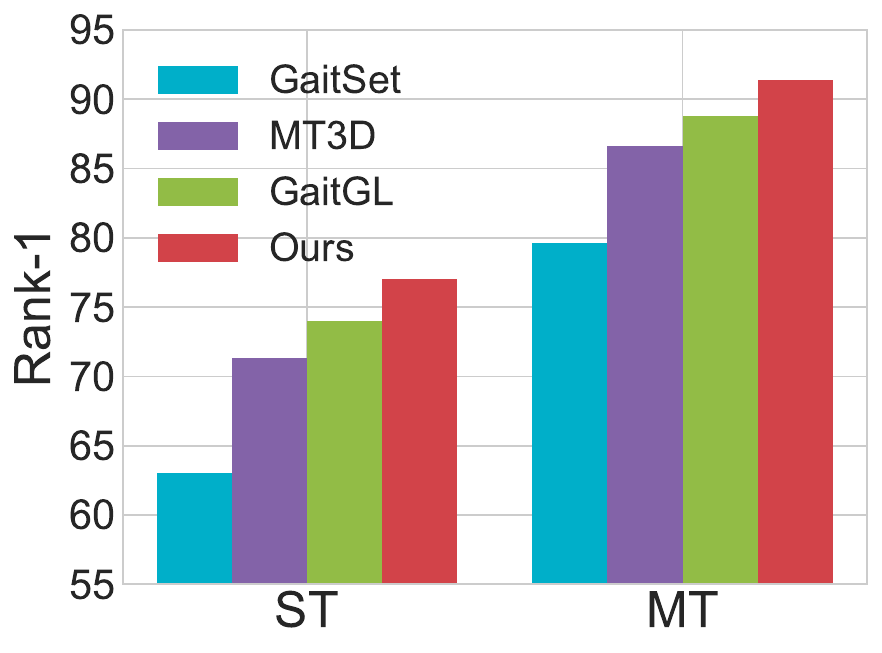}}	
\caption{Comparison under ST/MT setting~\cite{chao2019gaitset} on CASIA-B.}
\label{fig:stmt}
\end{figure}

\begin{table*}[t]
\renewcommand{\arraystretch}{0.8}
\setlength{\tabcolsep}{3mm}
\centering
\caption{Rank-1 (\%), Rank-5 (\%), mAP (\%), and mINP (\%) comparison on Gait3D. As skeleton-based methods are unrelated to the resolution, we only report one group of results. ``*'' denotes the method with extra 3D modality.}
\begin{tabular}{l|c|cccc|cccc}
\toprule
\multicolumn{2}{c|}{Input Size (H$\times$W)}  &\multicolumn{4}{c|}{128$\times$88}   & \multicolumn{4}{c}{64$\times$44} \\ \midrule
Methods                      			& Venue & Rank-1  & Rank-5  & mAP  & mINP & Rank-1 & Rank-5 & mAP  & mINP \\ \midrule
PoseGait~\cite{liao2020model}	        & PR20	& 0.2  & 1.1  & 0.5  & 0.3	& - & - & -  & - \\ 
GaitGraph~\cite{teepe2021gaitgraph}	& ICIP21& 6.3  & 16.2 & 5.2  & 2.4	& - & - & -  & - \\  \midrule
GaitSet~\cite{chao2019gaitset}		    & AAAI19	& 42.6 & 63.1 & 33.7 & 19.7	& 36.7 & 58.3 & 30.0 & 17.3	\\
GaitPart~\cite{Fan_2020_CVPR}		& CVPR20	& 29.9 & 50.6 & 23.3 & 13.2	& 28.2 & 47.6 & 21.6 & 12.4	\\ 
GLN~\cite{hou2020gait}				& ECCV20	& 42.2 & 64.5 & 33.1 & 19.6	& 31.4 & 52.9 & 24.7 & 13.6	\\
GaitGL~\cite{Lin_2021_ICCV}	            & ICCV21	& 23.5 & 38.5 & 16.4 & 9.2	& 29.7 & 48.5 & 22.3 & 13.3	\\ 
CSTL~\cite{Huang_2021_ICCV}	            & ICCV21	& 12.2 & 21.7 & 6.4  & 3.3	& 11.7 & 19.2 & 5.6  & 2.6	\\ \midrule

SMPLGait*~\cite{Zheng_2022_CVPR}& CVPR22  & 53.2	& 71.0 & 42.4 & 26.0 & 46.3& 64.5 & 37.2 & 22.2  \\  \midrule
\textbf{Ours}       & --  & \textbf{58.9}	& \textbf{75.7} & \textbf{47.3} & \textbf{29.4} & \textbf{52.4} & \textbf{69.2} & \textbf{40.2} & \textbf{24.9}  \\
\bottomrule
\end{tabular} 

\label{tab:gait3d}
\end{table*}

\begin{table}[t]
  \centering
  \renewcommand\arraystretch{0.8} 
  \setlength{\tabcolsep}{1.1mm}
  \caption{Rank-1 (\%), Rank-5 (\%), Rank-10 (\%), and Rank-20 (\%) performance comparison on GREW.}
\begin{tabular}{l|c|cccc}
\toprule
Method & Venue & Rank-1 & Rank-5 & Rank-10 & Rank-20 \\
\midrule
    PoseGait~\cite{liao2020model}& TPAMI16& 0.2  & 1.1  & 2.2  & 4.8  \\ 
    GaitGraph~\cite{teepe2021gaitgraph}& ICIP21& 1.3  & 3.5  & 5.1  & 7.5  \\
    \midrule
    GEINet~\cite{shiraga2016geinet} &ICB16 & 6.8   & 13.4  & 17.0  & 21.0      \\
    TS-CNN~\cite{Wu2017}& TPAMI16& 13.6  & 24.6  & 30.2  & 37.0  \\ \midrule
    GaitSet~\cite{chao2019gaitset} & AAAI19& 46.3  & 63.6  & 70.3  & 76.8  \\
    GaitPart~\cite{Fan_2020_CVPR}& CVPR20& 44.0  & 60.7  & 67.3  & 73.5  \\
    GaitGL \cite{Lin_2021_ICCV} &ICCV21 & 47.3  & 63.6  & 69.3  &74.2  \\ \midrule
    \textbf{Ours}  & --& \textbf{67.0} & \textbf{78.9} & \textbf{83.0} & \textbf{85.8} \\
       \bottomrule
\end{tabular}
\label{tab:grew}
\end{table}

\subsection{Comparison with State-of-the-art Methods}

\noindent\textbf{OU-MVLP.} To validate the generalizability, we conduct a comparison between the SOTA methods and CLASH on OU-MVLP in~\cref{tab:ou}. CLASH outperforms the previous methods by a considerable margin, proving the generalizability of CLASH in large dataset scenarios. Further, as DyGait~\cite{wang2023dygait} points out that extreme viewpoints, such as 0$^{\circ}$ and 180$^{\circ}$, contain less gait information, CLASH largely improves the performance under these viewpoints, which may indicate that dense gait descriptor could capture more subtle visual clues to extract walking pattern under these viewpoints.

\noindent\textbf{CASIA-B.} To validate the effectiveness of  CLASH on cross-view and cross-condition scenarios, we conduct a comparison between CLASH and state-of-the-art methods at the resolution of $64\times 44$ and $128\times 88$ as shown in~\cref{tab:avg} and~\cref{tab:avg128}, respectively. Although previous works achieve remarkable performance, CLASH still obtains a large performance gain at both two input resolutions. CLASH outperforms prevailing methods by 1.6\% and 1.4\% at the resolution $64\times 44$ /$128\times 88$ at least, respectively. Several conclusions can be summarized as follows. First, CLASH achieves rank-1 accuracy over 98\% and 96\% under NM and BG with subtle appearance differences, respectively. Second, as 3DLocal~\cite{Huang_2021_ICCV2} points out that the motion pattern dominates the recognition under CL, the performance of CLASH under CL condition achieves 89.3\%, which is attributed to the dense temporal information.

Then, we evaluate CLASH under more challenging protocols~\cite{chao2019gaitset} in~\cref{fig:stmt}, \textit{i.e.,} data-limited ST/MT setting~\cite{chao2019gaitset}. CLASH largely outperforms previous methods by a significant margin, exceeding GaitSet by 14.3\%/12.0\%, MT3D by 6.0\%/5.0\%, and GaitGL by 3.3\%/2.8\% on ST/MT setting, respectively. Further, CLASH under MT setting even achieves comparable performance with the latest SOTA  methods~\cite{Huang_2021_ICCV2, Huang_2021_ICCV, Lin_2021_ICCV} under LT setting, which further indicates the efficiency of CLASH.

\noindent\textbf{Gait3D.} We compare CLASH with prevailing methods on the latest in-the-wild dataset Gait3D in~\cref{tab:gait3d}. CLASH outperforms previous silhouette-based methods by 16.3\% and 15.7\% at the resolution of $128\times 88$ and $64\times 44$, respectively. Moreover, CLASH outperforms multi-modal SMPLGait by 5.7\% and 6.1\% at the resolution of $128\times 88$ and $64\times 44$, respectively. Note that SMPL requires extra 3D Mesh annotation while DSTF is transformed from silhouettes.

\noindent\textbf{GREW.} Further, we evaluate the proposed CLASH on in-the-wild GREW competition as shown in~\cref{tab:grew}. First, prevailing methods under the in-the-lab scenario suffer from a sharp performance decrease when transferred to the in-the-wild scenario.  Second, CLASH largely outperforms previous methods by 19.7\%, which indicates the necessity of dense representation of gait patterns and the effectiveness of complementary learning under in-the-wild scenarios.

\subsection{Ablation Study}

In this section, extensive ablation studies are conducted on the mainstream datasets to verify the effectiveness of each component of CLASH. Since NCL aims to extract the complementarity between two gait descriptors, the order of the ablation study is to verify the effectiveness of the DSTF first and the effectiveness of complementary learning using silhouette and the DSTF. Our baseline refers to the feature extractor with four 3D convolution layers only using silhouette. Then, to verify the validity of DSTF, we add Bi-DT and foreground/background separation successively, where the fusion strategy is simple element-wise add. Finally, to verify the validity of the NCL, we replace element-wise add with the searched architecture by NCL. Further, more qualitative and quantitative analyses are presented.

\begin{table}[h]
\centering
\setlength{\tabcolsep}{1.7mm}
\renewcommand{\arraystretch}{0.35}
\caption{Effectiveness of the components of CLASH, including Bi-DT, foreground/background separation strategy (F/B Sep.), and the neural architecture search based complementary learning (NCL).}
\begin{tabular}{ccc|cccc} \toprule
Bi-DT & F/B Sep. & NCL & CASIA-B & OU-MVLP & GREW & Gait3D \\ \midrule
          &          &     & 89.5   & 90.1   & 52.3   &  43.3    \\
\CheckmarkBold            &        &  &  91.4   & 90.3   & 60.9       & 47.9     \\
\CheckmarkBold        & \CheckmarkBold    &     & 91.8    & 91.3   & 62.5     & 48.9     \\
\CheckmarkBold & \CheckmarkBold         & \CheckmarkBold    & \textbf{93.9}   & \textbf{91.9}   &\textbf{67.0}    &   \textbf{52.4} \\ \bottomrule
\end{tabular}
\label{tab:each}
\end{table}
\begin{table}[t]
\centering
\renewcommand{\arraystretch}{0.85}
\caption{Comparison between NCL and other operations.}
\begin{tabular}{l|cccc}
\toprule
Method            & CASIA-B  & OU-MVLP & GREW & Gait3D \\ \midrule
Sum            &91.8     &  91.3  & 62.5   &48.9   \\
Concat           &  92.0   & 91.3   & 62.7   & 49.0   \\
Conv                & 92.7   &  91.5  & 63.5   & 49.6    \\
Cross-attention    &  92.5   & 91.5   & 64.3   &  51.0   \\
Channel attention  &  93.0  &  91.6  & 65.1   &  51.2   \\
MMGaitFormer~\cite{Cui_2023_CVPR}  &  92.9 &  91.7  & 64.8   &  51.5   \\
NCL              &  \textbf{93.9}   & \textbf{91.9}    &\textbf{67.0}      &  \textbf{52.4}    \\ \bottomrule
\end{tabular}
\label{tab:ncl}
\end{table}
\begin{figure*}[ht]
\centering
	\subcaptionbox{NM}{\includegraphics[width = 0.25\textwidth]{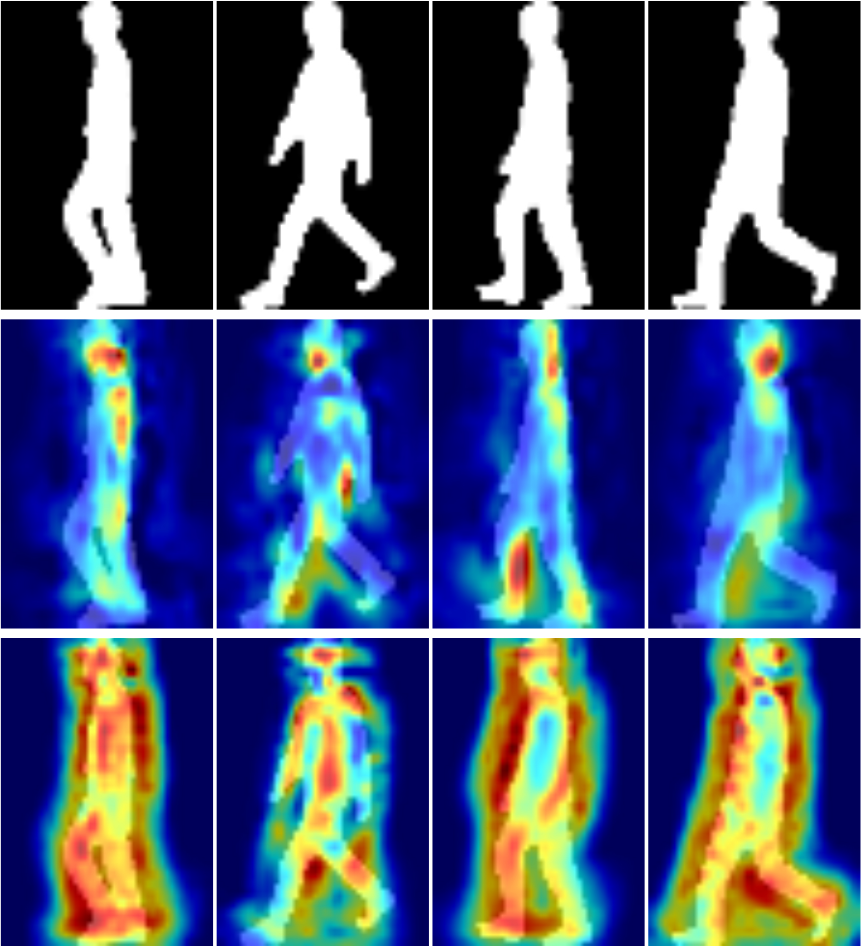}}
	\hspace{0.5cm}
	\subcaptionbox{BG}{\includegraphics[width = 0.25\textwidth]{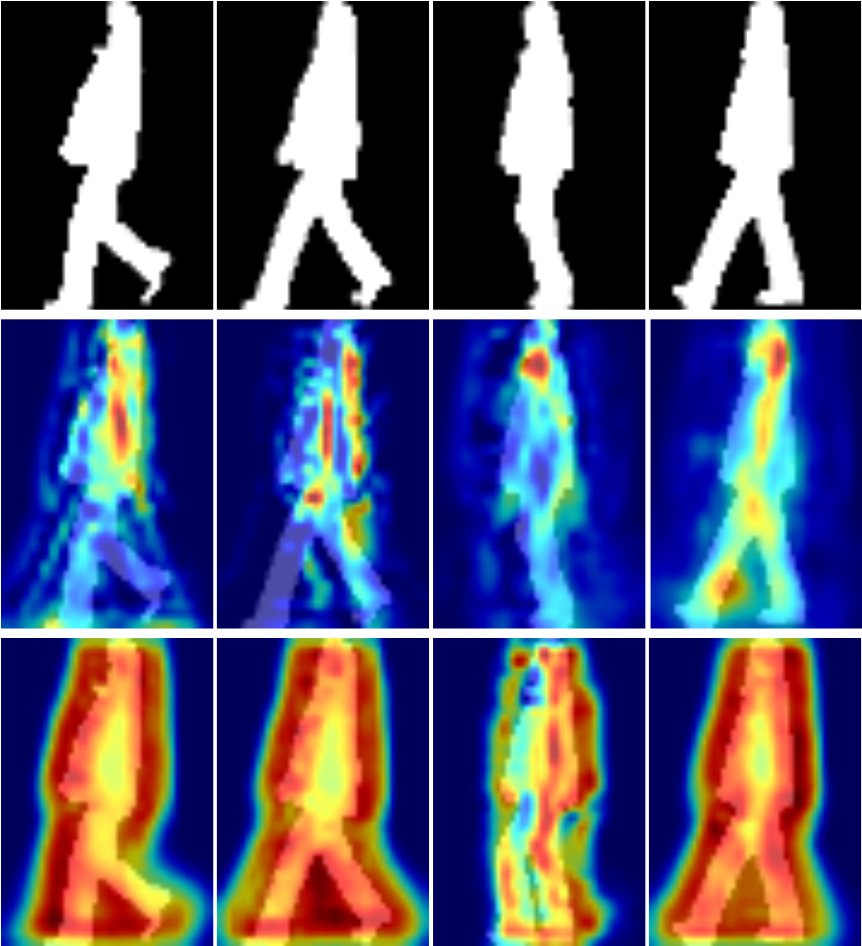}}
	\hspace{0.5cm}
	\subcaptionbox{CL}{\includegraphics[width = 0.25\textwidth]{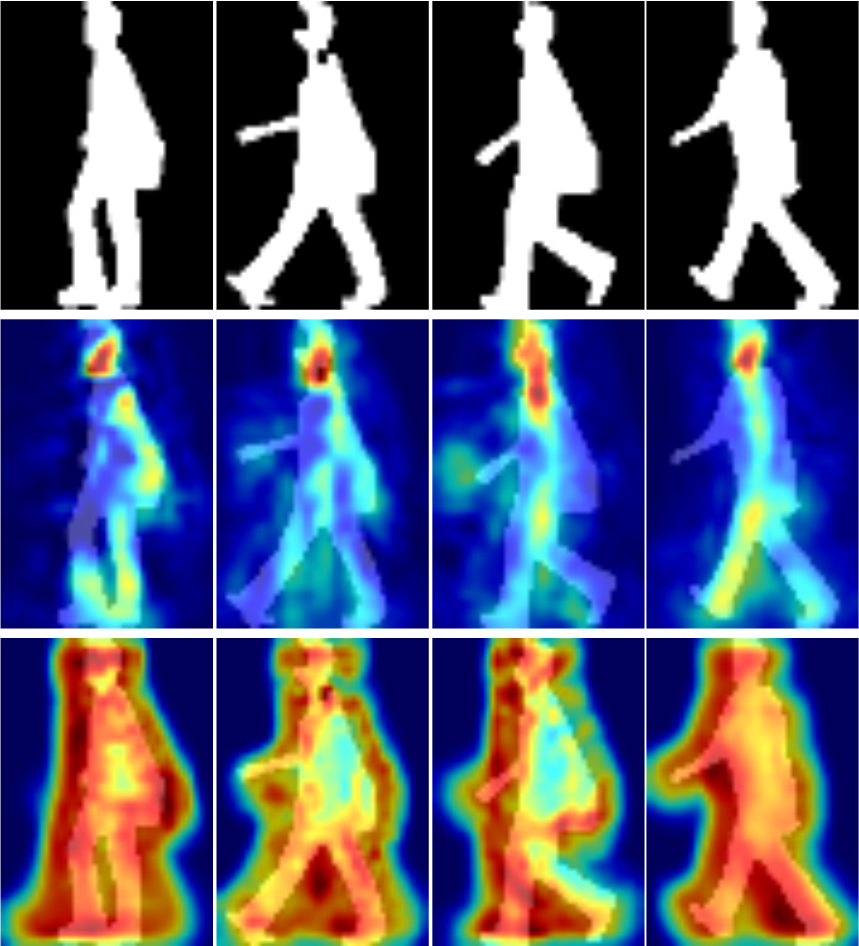}}
\caption{Grad-CAM visualization under NM, BG, and CL conditions. The Grad-CAM of three rows refers to silhouette, Grad-CAM with baseline, and Grad-CAM with CLASH, respectively.}
\label{fig:gradcam}
\end{figure*}

\noindent\textbf{Effectiveness of DSTF and NCL.} To validate the effectiveness of DSTF and NCL, we conduct an ablation study on them individually. The first row in~\cref{tab:each} is the baseline. Considering NCL is performed on two descriptors, the order of the ablation is DSTF first and then NCL. Note that DSTF is composed of Bi-DT and foreground/background separation. The individual impacts of DSTF and NCL are shown in~\cref{tab:each}. The 2.3\% and 2.1\% performance gains from DSTF and NCL indicate the effectiveness of dense spatial-temporal field and NAS-based complementary learning, respectively.

\noindent\textbf{Analysis of DSTF.} We analyze DSTF from its Bi-DT and foreground/background separation strategy (F/B Sep.) in~\cref{tab:each}, respectively. First, improving the sensitivity to the walking pattern with distance-based texture based representation (Bi-DT)  could improve rank-1 performance by 1.9\%. Second, the 0.4\% performance gain brought by the foreground/background separation strategy indicates its necessity and effectiveness. 
\begin{table}[t]
\caption{Comparison of regularization impact between common regularization methods and silhouette. The complementary learning layer between DSTF and silhouette is the element-wise sum.The first row denotes the baseline but replaces silhouette with DSTF.}
\centering
\renewcommand{\arraystretch}{0.85}
\begin{tabular}{l|cccc}
\toprule
Method  & CASIA-B & OU-MVLP & GREW & Gait3D\\ \midrule
DSTF     &  90.3     & 90.7   & 57.8   & 46.5 \\
\hspace{1em}+ Label Smooth       &  90.6    &   90.9  & 59.3    & 47.0  \\
\hspace{1em}+ Dropout       &  90.9  &  91.0   & 61.2   & 47.5    \\
\hspace{1em}+ $L_2$ Norm           &  90.7   & 91.0   & 60.5   & 47.2    \\
  \hspace{1em}+ Silhouette           &\textbf{91.8}   & \textbf{91.3}   &\textbf{62.5}    &\textbf{48.9}      \\ \bottomrule
\end{tabular}
\label{tab:regu}
\end{table}
\begin{table}[t]
\centering
\setlength{\tabcolsep}{2.8mm}
\renewcommand{\arraystretch}{0.8}
\caption{Generalizability of CLASH on prevailing  methods. $\dagger$ denotes the method equipped with DSTF and NCL.}
\begin{tabular}{c|cccc}
\toprule
Method   &CASIA-B & OU-MVLP& GREW & Gait3D \\ \midrule
GaitSet   & 84.2 &  87.1 & 46.3 & 36.7    \\
GaitSet$\dagger$    & 89.0  & 89.5   & 61.2   & 55.9    \\

GaitPart  & 88.8  &   88.7 & 44.0 & 28.2\\
GaitPart$\dagger$   & 91.3 &  90.6 &60.3   & 46.9     \\
GaitGL   & 91.8   &  89.7 & 47.3 & 29.7    \\
GaitGL$\dagger$   &  94.3     &  92.0  & 65.2   &  48.8\\ 
GaitBase~\cite{Fan_2023_CVPR}   & 89.7     &   90.0 & 60.1 & 65.6  \\
GaitBase$\dagger$    &  94.1     & 92.2   &  71.2   & 73.0   \\
\bottomrule
\end{tabular}
\label{tab:plus}
\end{table}

\noindent\textbf{Analysis of Complementary Learning.} We analyze the effectiveness of complementary learning by comparing the frequently-used operations with the searched MD cell by NCL in~\cref{tab:ncl}, where Conv is a 3D CNN with approximately equal parameters as the searched MD cell for fairness. First, frequently-used operations (\textit{e.g.}, sum and concat) could effectively improve the performance compared to using silhouette alone, indicating the necessity of complementary learning. Second, MD cell searched by NCL could significantly improve the performance by 2.1\%, which indicates that NCL could efficiently search for an effective complementary learning architecture to outperform manually designed operations efficiently.

\noindent\textbf{Regularization Impact of Silhouette.} Since sparse representation could be a regularization method~\cite{li2017weighted, shao2018spatial}, we compare the regularization impact of the silhouette with several regularization methods in~\cref{tab:regu}, which may indicate that sparse boundary representation has superior regularization impact on dense texture representation. Besides, although the performance of DSTF alone is already 0.8\% higher than that of silhouette alone, the regularization of silhouette could further unlock the potential of DSTF.

\noindent\textbf{Generalizability of DSTF and NCL.} To validate the generalizability of the proposed methods, we replace the feature extractor with prevailing methods. The comparison shown in~\cref{tab:plus} indicates the effectiveness and generalizability of DSTF and NCL. Further, it also shows the potential of silhouette-based methods with dense representation modeling.

\begin{table}[h]
\centering
\setlength{\tabcolsep}{1.7mm}
\renewcommand{\arraystretch}{0.4}
\caption{Ablation on the selection of gait descriptor in complementary learning.}

\begin{tabular}{ccc|cccc} \toprule
 Silhouette        & DSTF & GEI & CASIA-B & OU-MVLP & GREW & Gait3D \\ \midrule
\CheckmarkBold &  \CheckmarkBold&  &   93.9       & 91.9        &   67.0   &    52.4    \\
\CheckmarkBold&  & \CheckmarkBold &   90.6       &    90.5      &   54.7   & 45.3     \\ 
& \CheckmarkBold&\CheckmarkBold &92.7&91.2&62.1& 49.8\\ \bottomrule
\end{tabular}
\end{table}

\noindent\textbf{Selection of gait descriptor} To validate the effectiveness of DSTF compared to other gait descriptors with complementary learning, we compare the complementary learning using different gait descriptors. The extra gait descriptor is adopted as GEI~\cite{1561189}. The results indicate that complementary learning between silhouette and DSTF is more effective than that of silhouette and GEI. Besides, complementary learning of DSTF and GEI could improve the performance compared to baseline but is inferior to silhouette and DSTF.

\noindent\textbf{Network Attention Visualization}
To validate the effectiveness of CLASH intuitively,  we apply Grad-CAM~\cite{8237336} to the baseline and proposed CLASH for the qualitative analysis in~\cref{fig:gradcam}. Grad-CAM is used to identify the regions that the network considers essential based on the gradient response, \textit{i.e.,} network attention.

For the baseline, there is no stable attention pattern since it may sparsely focus on the different local parts of the whole silhouette. Therefore, the baseline model may tend to be affected by the covariates. By contrast, the attention pattern of CLASH is stable to focus on more densely distributed attention on the human body than baseline, which indicates CLASH could learn an informative and robust representation to be less dependent on the covariate.
 \begin{figure}[t]
 \small
    \begin{center}
       \includegraphics[width=0.4\textwidth]{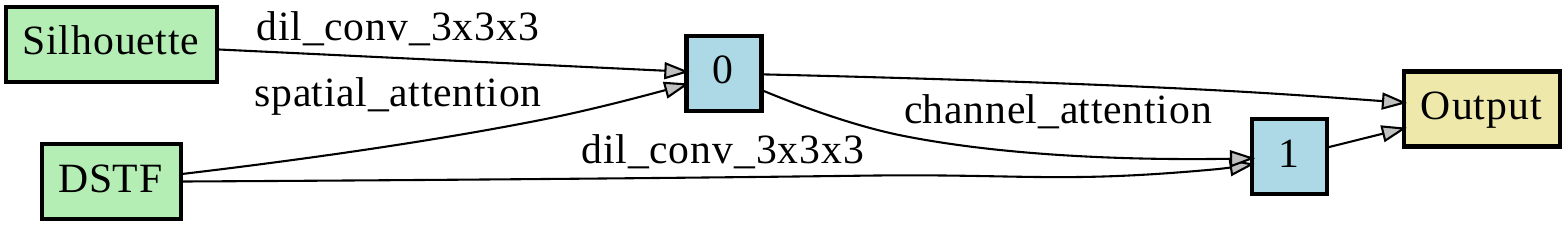}
    \end{center}
    \caption{Searched architecture of MD cell.}
    \label{fig:searched}
 \end{figure}

\noindent\textbf{Searched Architecture of MD Cell.}
The searched architecture of the MD cell is shown in~\cref{fig:searched}. Complementary learning between two heterogeneous descriptors requires more complex operations, \textit{i.e.,} spatial attention and channel attention, and the NAS-based complementary learning could avoid inefficient trial and error. Further, the searched architecture inspires us to expand the receptive field and follow the spatial-then-channel attention mechanism.

\begin{figure}
\centering
	\subcaptionbox{Baseline}{\includegraphics[width = 0.15\textwidth]{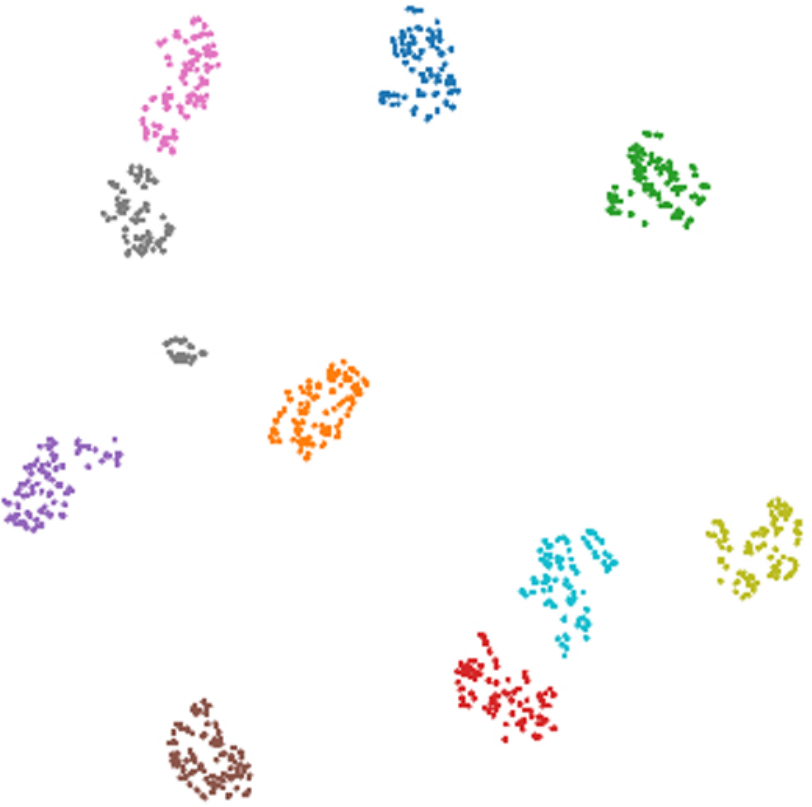}}
	\hspace{1.3cm}
	\subcaptionbox{Ours}{\includegraphics[width = 0.15\textwidth]{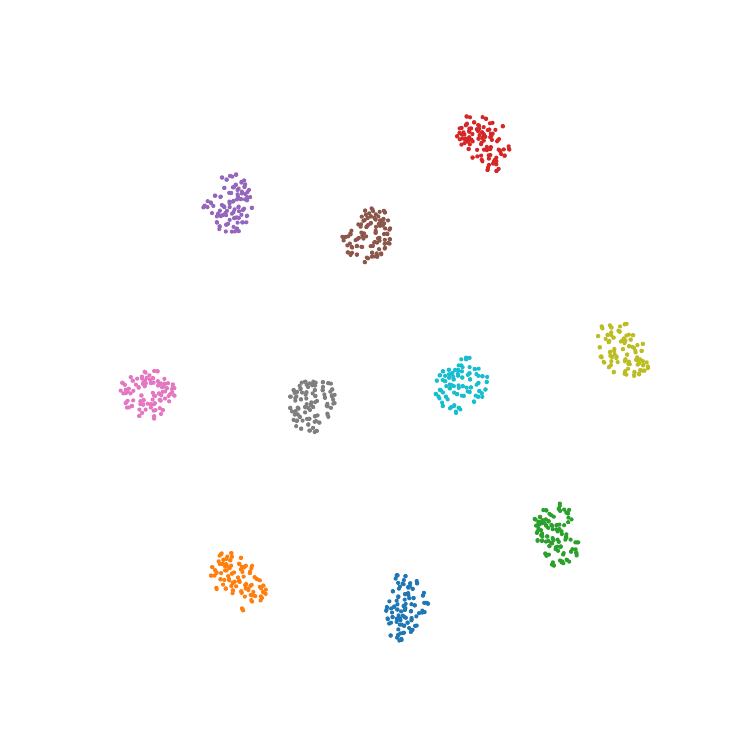}}
	\hfill
\caption{Comparison of the feature space.}
\label{fig:vis}
\end{figure}

\noindent\textbf{The Visualization of Feature Space.}  We randomly choose 10 subjects from CASIA-B to visualize their feature distributions using t-SNE~\cite{van2008visualizing}. Comparing the feature distributions of baseline and proposed CLASH in~\cref{fig:vis}, we find that CLASH effectively improves the intra-class compactness and inter-class separability,  which proves the discriminative feature representation ability of CLASH.

\label{sec:exp}

\section{Conclusion}
In this paper, we propose complementary learning with neural architecture search (CLASH) framework, consisting of walking pattern sensitive gait descriptor named dense spatial-temporal field (DSTF) and neural architecture search based complementary learning (NCL) method. Specifically, DSTF models the spatial-temporal context information by transforming the representation from the sparse binary boundary into the dense distance-based texture. NCL mutually complements the sensitivity of texture-based representation and the robustness of boundary-based representation through efficient NAS-based complementary learning. Extensive experiments on public datasets verify the effectiveness of our method under both in-the-lab and in-the-wild scenarios.

\label{sec:conclusion}

\section*{Acknowledgements}
This work is supported in part by National Natural Science Foundation of China under Grant U20A20222, National Science Foundation for Distinguished Young Scholars under Grant 62225605, Zhejiang Key Research and Development Program under Grant 2023C03196,  and sponsored by CCF-AFSG Research Fund, CCF-Zhipu AI Large Model Fund (CCF-Zhipu202302) as well as The Ng Teng Fong Charitable Foundation in the form of ZJU-SUTD IDEA Grant, 188170-11102. This work is also supported by Ant Group.

\normalem
\bibliography{sample}
\bibliographystyle{IEEEtran}

\begin{IEEEbiography}[{\includegraphics[width=1in,height=1.25in,clip,keepaspectratio]{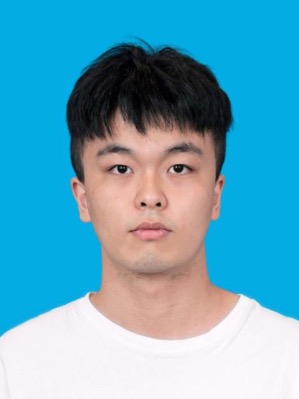}}]
  {Huanzhang Dou} received the bachelor's degree in computer science and technology from Northeastern University, China, in 2020.
  He is currently pursuing the Ph.D. degree with the College of Computer Science, Zhejiang University, Hangzhou, China, under the supervision of Prof. X. Li.
  His current research interests are primarily in video diffusion models, multi-modal learning, and biometrics.
\end{IEEEbiography}
\vspace{-1.5cm}
\begin{IEEEbiography}[{\includegraphics[width=1in,height=1.25in,clip,keepaspectratio]{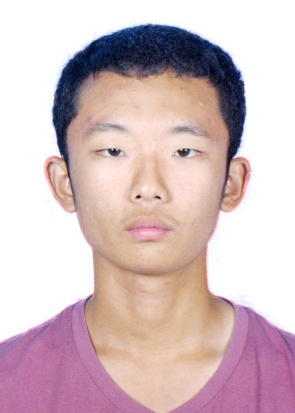}}]
  {Pengyi Zhang} received the bachelor's degree in computer science and technology from Beijing Institute of Technology, China, in 2020. He is currently pursuing the master's degree with the College of Computer Science, Zhejiang University, Hangzhou, China, under the supervision of Prof. X. Li. His current research interests are primarily in computer vision and deep learning.
\end{IEEEbiography}
\vspace{-1.5cm}
\begin{IEEEbiography}[{\includegraphics[width=1in,height=1.25in,clip,keepaspectratio]{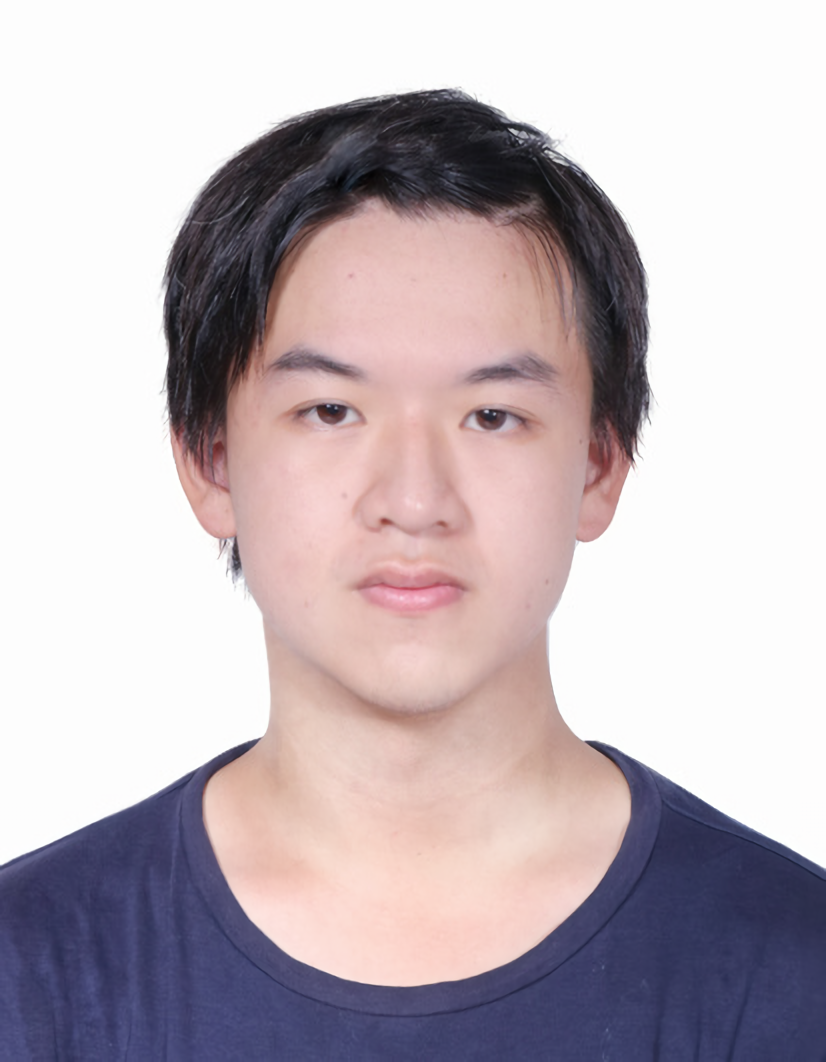}}]
  {Yuhan Zhao} received the bachelor’s degree in Software engineering from Sichuan University, China, in 2020. He is currently pursuing the master’s degree with the College of Software Technology, Zhejiang University, Hangzhou, China, under the supervision of Prof. X. Li. His current research interests are primarily in computer vision and deep learning.
\end{IEEEbiography}
\vspace{-1.5cm}
\begin{IEEEbiography}[{\includegraphics[width=1in,height=1.25in,clip,keepaspectratio]{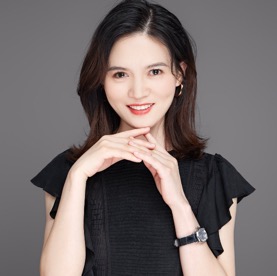}}]
  {Lu Jin} is a senior security expert. Head of End Confrontation in Ant Group‘s Cybersecurity Department, responsible for the Ant face security and the construction of the end-wide threat confrontation system.
\end{IEEEbiography}
\vspace{-1.5cm}
\begin{IEEEbiography}[{\includegraphics[width=1in,height=1.25in,clip,keepaspectratio]{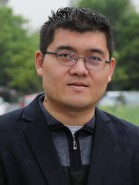}}]
 {Xi Li} received the Ph.D. degree from the National Laboratory of Pattern Recognition, Chinese Academy of Sciences, Beijing, China, in 2009. 
 From 2009 to 2010, he was a Post-Doctoral Researcher with CNRS, Telecomd ParisTech, France. He is  currently a Full Professor with Zhejiang University, China. Prior to that, he was a Senior Researcher with the University of Adelaide, Australia. His research interests include artificial intelligence, computer vision, and machine learning.
\end{IEEEbiography}

\newpage

\vspace{11pt}

\vfill

\end{document}